\documentclass[letterpaper]{article}
\usepackage{natbib,alifeconf}
\usepackage{bm}
\pdfoutput=1

\graphicspath{ {./images/} }

\title{Ladder Networks for Semi-Supervised Hyperspectral Image Classification}
\author{Julian B{\"u}chel$^{1,2}$ \and Okan Ersoy$^{1}$ \\
\mbox{}\\
$^1$Purdue University, 610 Purdue Mall, West Lafayette, IN, USA \\
$^2$ETH Zürich, R{\"a}mistrasse 101, Z{\"u}rich, CH \\
jubueche@ethz.ch}

\begin{document}
\maketitle

\begin{abstract}
We used the Ladder Network [\cite{DBLP:journals/corr/RasmusVHBR15}] to perform Hyperspectral Image Classification in a semi-supervised setting. The Ladder Network distinguishes itself from other semi-supervised methods by jointly optimizing a supervised and unsupervised cost. In many settings this has proven to be more successful than other semi-supervised techniques, such as pretraining using unlabeled data. We furthermore show that the convolutional Ladder Network outperforms most of the current techniques used in hyperspectral image classification and achieves new state-of-the-art performance on the Pavia University dataset given only 5 labeled data points per class.
\end{abstract}

\section{Introduction}
Over the last years, semi-supervised learning has increasingly gained interest in the Machine Learning community. Having only a few labeled data points has become the reality in many real-world applications. This is due to the fact that labeling data points is tedious and sometimes costly work. \newline

This problem has also gotten attention in the Hyperspectral Imaging community and various techniques have been proposed in order to solve it [\cite{8071701}, \cite{7729675}, \cite{7729473}, \cite{8071798}, \cite{7326799}, \cite{8105856}, \cite{8451525}, \cite{8517816}, \cite{7936545}, \cite{7985433}, \cite{8518846}]. A lot of those techniques make use of the unlabeled training data by pretraining in an unsupervised setting and afterwards performing supervised learning on the labeled examples. One example is the use of a denoising Auto-Encoder[\cite{Vincent2010StackedDA}, \cite{7405235}] as the unsupervised learning technique and a simple convolutional neural network as the supervised learning mechanism. The Auto-Encoders goal is to reconstruct the original input after various transformations into a lower dimensional space. The learned representation can then be used for supervised training. The question arises whether the learned representation is valuable for the supervised learning task or not. Often it is only partially useful. The Auto-Encoder retains information useful for reconstruction and not necessarily useful for classification. The Ladder Network (LN) resolves this problem by \textbf{jointly} optimizing a reconstruction and classification cost. It should be noted that \cite{doi:10.1080/2150704X.2017.1331053} have used the Ladder Network in hyperspectral image classification, but we achieved even better accuracy on 5 times less labeled training examples per class. Our paper further investigates the effect of different components in the Ladder Network and presents a comparison to existing state-of-the-art algorithms. We additionally show that the Ladder Network achieves the best performance given only 5 labeled examples per class.\newline

In the following section, we will provide an overview of how the Ladder Network works and what its key components are. In the experimental section, we compare two variants of the Ladder Network: One based on simple, fully-connected layers and the other one based on convolutional layers. We show that for any number of labeled training examples, the convolutional LN outperforms the fully connected LN. We then compare the two networks with other competitive techniques on two widely used datasets in the Hyperspectral Imaging community and show that Ladder Networks are superior to most of the current models.  \newline

At the end, we conclude our findings and describe further possibilities on how to gain higher accuracies using this approach.

\section{The Ladder Network}
The Ladder Network has a complex architecture, which can be derived by systematically adding more features to simpler architectures. For this reason, we will start with a simple Auto-Encoder and gradually build upon it to derive the Ladder Network. \newline

Auto-Encoders (AE) consist of two parts: The \textit{Encoder} and \textit{Decoder}. The encoder maps input $\bm{x}$, using a function $\bm{f_{\theta}(x)}$, to an output $\bm{y}$, which has a lower dimension. The decoder on the other hand maps the input $\bm{y}$ to the output $\bm{z}$ using another transformation $\bm{g_{\theta'}(y)}$. The output $\bm{z}$ is the reconstructed input $\bm{x}$.  The parameters $\bm{\theta}$ and $\bm{\theta'}$ of the transformations are learned via backpropagation by minimizing the reconstruction cost between $\bm{x}$ and $\bm{z}$. \newline

There are two fundamental issues with traditional AE's: First, the one-layer architecture may be too simple to learn complex features and second the network may just learn some kind of identity mapping. \newline
The second problem can be resolved by simply reducing the dimensionality or by adding random noise. Adding noise has the effect that parts of the input are perturbed and the AE therefore needs to extract relevant information from the input by denoising it. Hence the name Denoising Auto-Encoder (\textbf{DAE}). In order to overcome the first problem, it is natural to think of simply adding more layers: Stacked Denoising Autoencoders \textbf{(SDAE)} are single \textbf{DAEs}, which try to extract features from the encoded input of a previous \textbf{DAE}. Figure 1a illustrates the unsupervised training process. We can then use the encoded input as a feature for any given classification task as shown in Figure 1b.

\begin{figure}[!htb]
\begin{center}
\includegraphics[width=3in]{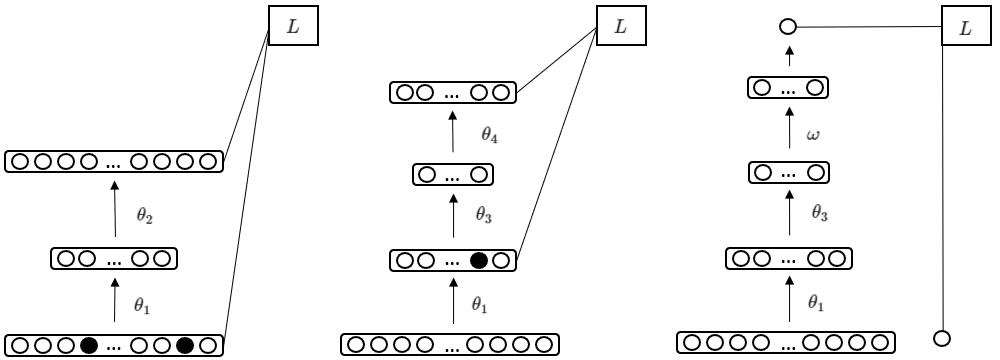}
\caption{Figure 1a (left, middle) shows the unsupervised training process of a \textbf{SDAE} of size 2. Figure 1b (right) shows the network of stacked encoders with an additional layer performing a classification task. All the weights are subject to change during training of the complete network (right). Black dots symbolize data points that have been zeroed out.}
\label{fig1}
\end{center}
\end{figure}

We have provided a GoogleCollaborator notebook with implementations of the various Auto-Encoders, which can be found in the section 'Code'. \newline

The \textbf{Ladder Network} is essentially a stacked denoising Auto-Encoder with lateral skip connections. It combines the cost of a supervised task with the reconstruction cost of intermediate representations of the input. Figure 2[\cite{DBLP:journals/corr/RasmusVHBR15}] illustrates the architecture.

\begin{figure}[!htb]
\begin{center}
\includegraphics[width=3in]{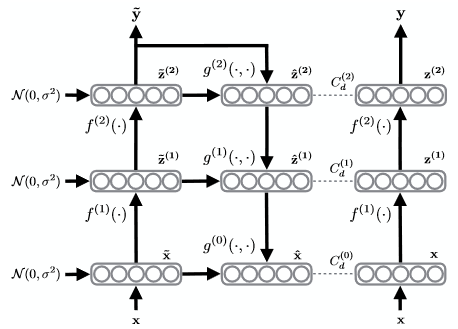}
\caption{During training, the input $\bm{x}$ is corrupted and transformed in the encoder (left). The supervised training is done using the corrupted representation of $\bm{x}$. The corrupted input is then reconstructed (middle) using the decoder. As targets for the reconstruction, the clean encoder representations are used (right).}
\label{fig2}
\end{center}
\end{figure}

Initially, the network performs a noisy pass through the encoder. Input $\bm{x}$ is corrupted using Gaussian noise with zero mean and a user defined standard deviation. The intermediate layers perform linear or non-linear transformations of the input followed by batch normalization. Precisely, $\bm{x}$ initially gets transformed to \[ \bm{ \tilde{h}^{(0)} = x+\textbf{noise}}\] and for all layers $\bm{l}$, \[ \bm{\tilde{z}^{(l)} = \textbf{batchnorm}(W^{(l)}\tilde{h}^{(l-1)})}+\textbf{noise} \]
Using any activation (for example ReLU) we obtain
\[ \bm{ \tilde{h}^{(l)} = \textbf{activation}(    \gamma^{(l)} *  (\tilde{z}^{(l)} + \beta^{(l)})  )  }\]
and repeat the process.
Afterwards, the most abstract representation of the input, namely $\bm{  \tilde{h}^{(L)} }$, is passed through the decoder. Initially, we have
\[ \bm{u^{(L)} = \textbf{batchnorm}(h^{(L)})}  \]
And subsequently for all layers $\bm{l}$
\[ \bm{   u^{(l)} = \textbf{batchnorm}(V^{(l+1)} \hat{z}^{(l+1)}  )   }  \]
where \[ \bm{\hat{z}^{(l+1)}  = g( \tilde{z}^{(l+1)},u^{(l+1)}  )}\]
Here, $\bm{g(.)}$ refers to the pointwise function, applied as the lateral skip connection. As shown in Figure 2 (right), we now pass the clean $\bm{x}$ through the encoder to obtain the intermediate representations denoted $\bm{z^{(l)}}$. The reconstruction cost is therefore the sum over all differences between original representation and reconstruction
\[  \bm{  \textbf{C}_{\textit{Recon}} = \sum_{l}{\lambda_l || z^{(l)}-\hat{z}^{(l)} ||^2  } }\]
where $\bm{\lambda_l}$ is a user defined multiplier.

The supervised cost is defined as
\[  \bm{ \textbf{C}_{\textit{Super}} = - \frac{1}{N}\sum_{i}{\textit{log}(P(\tilde{y_i} = t_i | x_i)) }  }\]
where $\bm{\tilde{y}}$ is the resulting prediction using the corrupted encoder pass. For prediction, we are using the clean encoder. The motivation behind using the corrupted encoder for the supervised training is the same as for the denoising Auto-Encoder: We are learning a denoising function, which essentially serves as a feature extractor. \newline
The total cost is
\[\bm{ \textbf{C}_{\textit{Total}} =  \textbf{C}_{\textit{Recon}} + \textbf{C}_{\textit{Super}} }\]
Several papers [\cite{7729856}, \cite{8278975}, \cite{8127330}] have exploited the spatial information carried in hyperspectral images by using Convolutional Neural Networks. Unfortunately, these networks lack the ability to generalize well given only few labeled data points, as we show in our experiments. By substituting ordinary, fully-connected layers in the Ladder Network by convolutional ones (Figure 3), we combine the benefits of CNN's and LN's.

\begin{figure}[!htb]
\begin{center}
\includegraphics[width=3.5in]{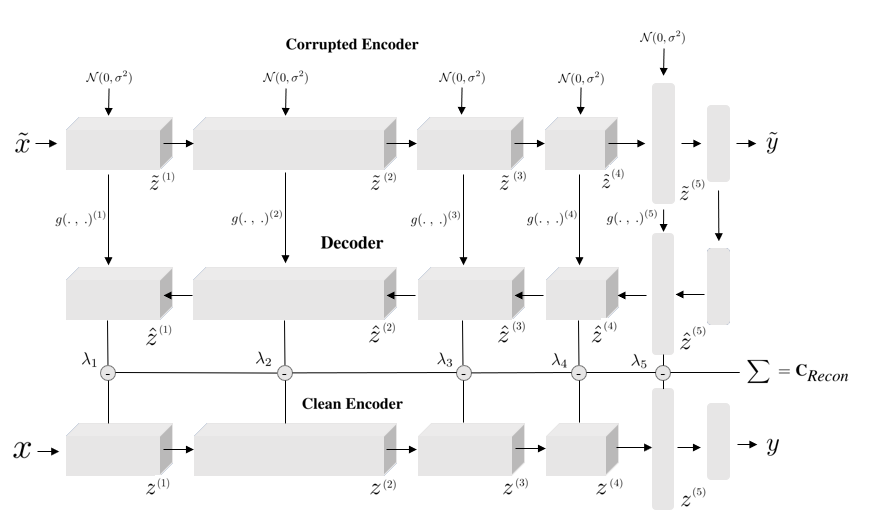}
\caption{The convolutional version of the Ladder Network.}
\label{fig3}
\end{center}
\end{figure}

\section{Experiments}
In this section, we compare the convolutional and fully-connected Ladder Network to various other methods on a widely used dataset in the Hyperspectral Imaging community. We show that the convolutional Ladder Network is able to achieve highly competitive results, given all labeled data points, but also is superior to most other techniques in the semi-supervised setting. The convolutional LN was able to achieve new state-of-the-art performance given 5 labeled points per class. We further analyze the effect of the denoising cost given different amounts of labeled data, the influence of the injected noise and the effect of dimensionality reduction using Principle Component Analysis (PCA) as a preprocessing step.
All the experiments were conducted on GoogleCollaborator and we have made the code available (see section 'Code').\newline

The dataset we used is a hyperspectral image of the Pavia University in Italy. This image was recorded using the ROSIS sensor, and after the removal of 12 bands due to noise and water absorption comprises 103 spectral bands, with a range of  0.41 to 0.82 $\mu m$. The 610x340 image and its corresponding groundtruth can be seen in Figure 4. Due to imbalance in the dataset, we either down- or upsampled the minority labels, depending on the method. In all experiments, we used 25\% of the points for testing and the remaining points for training. The code for the convolutional Ladder Network was written in TensorFlow[\cite{tensorflow2015-whitepaper}] and we used the Adam optimizer[\cite{DBLP:journals/corr/KingmaB14}] to train the network.

\begin{figure}[!htb]
\begin{center}
\includegraphics[width=3.5in]{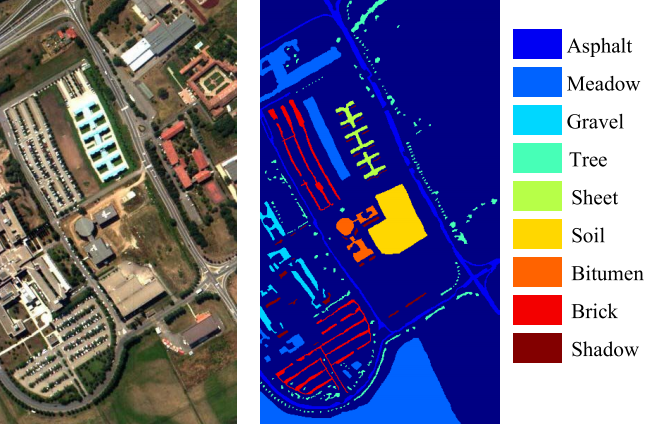}
\caption{The false color image (left) and the groundtruth image (right) of the Pavia University with the corresponding 9 classes.}
\label{fig4}
\end{center}
\end{figure}

Figure 5 shows the accuracy of the convolutional and fully-connected Ladder Network given different numbers of labeled training points. As can be seen, the fully-connected Ladder Network is constantly being outperformed by the convolutional version. For the fully-connected LN, we used a learning rate of 0.005, a batch size of 100, 4 intermediate layers with sizes 300-200-100-100 and respective denoising costs of 100-10-1-0.1-0.1. For the convolutional LN, we used a learning rate of 0.01, a batch size of 100 and the architecture was conv-90 - conv30 - conv15 - fc30, where the number after conv denotes the filter size. The window size achieving the best accuracy was 7. After each convolutional layer we used the ReLU activation function. The denoising cost for all layers that achieved the best accuracy was 0-0-0-0-0-0.42. 

\begin{figure}[!htb]
\begin{center}
\includegraphics[width=3.5in]{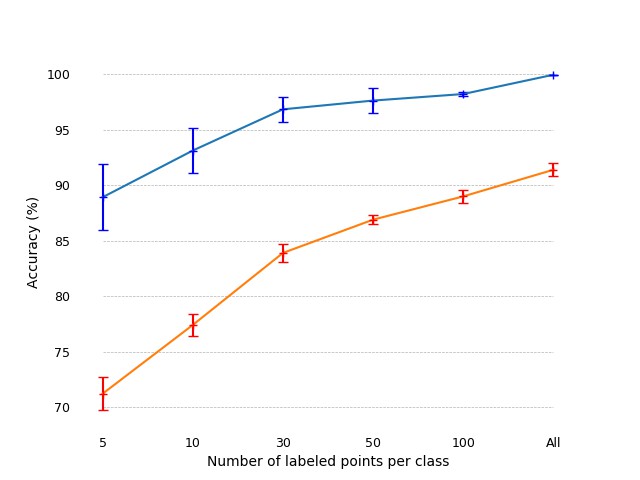}
\caption{Accuracy versus different numbers of labeled examples per class using the convolutional LN (blue) and the fully-connected LN (red).}
\label{fig5}
\end{center}
\end{figure}

\begin{table}
\begin{tabular}{lll}
Algorithm     & \multicolumn{2}{l}{Number of labeled points} \\ \hline
                  & 5 & 10 \\ \hline
CNN           &  55.40 $\pm$ 3.89 &  69.02 $\pm$ 2.26  \\
CDL-MD-L      &  72.85 &  82.61 $\pm$ 2.95  \\
Co-DC-CNN     & 83.47 $\pm$ 3.01  &  94.99 $\pm$ 1.49  \\
PNGrow        &  88.11 $\pm$ 2.87 &  93.85 $\pm$ 2.23  \\
TT-AL-MSH-MKE &  79.04 $\pm$ 3.95 &  86.00 $\pm$ 3.04 \\
$\textrm{S}^2$CoTraC      &  50.76 $\pm$ 1.68 &  80.75 $\pm$ 0.35  \\
Co-DC-Res & 86.69 $\pm$ 2.94 & $\bm{96.16 \pm 1.05}$ \\
FC-Ladder     &  71.2 $\pm$ 1.5  &  77.4 $\pm$ 1.0  \\
CNN-Ladder    & $\bm{88.92 \pm 2.97}$  &  93.13 $\pm$ 2.03 
\end{tabular}
\caption{Overall accuracies of different algorithms compared to the FC-LN and the CNN-LN.}
\label{table:tab1}
\end{table}

Table 1 illustrates that the convolutional LN achieved new state-of-the-art performance given only 5 labeled datapoints. During our experiments, we observed that the convolutional Ladder Network either converges to an overall accuracy of $\approx 96\%$ or $\approx 90\%$ and that the chance of converging to $96\%$ naturally increases with the number of labeled points. This shows that some points do carry more information than others and it would be interesting to try filter out the points not carrying any valuable information in order to guarantee convergence.

The algorithms we compared the Ladder Networks to are: \textbf{CNN}[\cite{7514991}], \textbf{CDL-MD-L}[\cite{MA201699}], \textbf{Co-DC-CNN}[\cite{doi:10.1080/2150704X.2017.1280200}], \textbf{PNGrow}[\cite{ROMASZEWSKI201660}], \textbf{TT-AL-MSH-MKE}[\cite{rs8090749}], $\textbf{S}^2$\textbf{CoTraC}[\cite{APPICE2017229}] and \textbf{Co-DC-Res}[\cite{codcres}]. The CNN is a basic deep CNN that uses spectral-spatial features to classify HSI images. CDL-MD-L is a contextual deep learning algorithm for the semi-supervised setting. It uses multi-decision labeling and is based on self-training. PNGrow is a semi-supervised algorithm that has two experts, namely the P and N-expert, which learn features from the spatial and spectral structure of the images respectively. PNGrow uses a co-training approach, just like Co-DC-CNN and Co-DC-Res, which employ a dual-channel (DC) strategy and use CNN's and ResNets as their building blocks. TT-AL-MSH-MKE is an active-learning based approach using a tri-training method for spectral-spatial learning. It furthermore uses a combination of MLR,KNN and ELM (MKE). $\textrm{S}^2$CoTraC is a semi-supervised, co-training based algorithm that extracts information at pixel level using collective inference. It should be noted that the overall accuracies reported were recorded by \cite{codcres}.
It is apparent that Co-DC-Res outperforms all other methods in the case of 10 labeled points per class, so it is worth explaining it in a little bit more detail: The algorithm proposed by Fang et al. consists of two main iterations. In the first iteration, the labeled points are used to train two ResNets, which are using the spectral and spatial information respectively. Both classifiers are then used to classify the unlabeled points and a sample selection strategy is used to select the most confident points for training the ResNets again. The employed sample selection strategy is based on the similarity of two given points. For the similarity metric, Fen et al. used a hierarchical representation of the input $x$, namely $[r_1,r_2,r_3]$, which are the output of the first convolutional layer and the two building blocks, respectively.

The striking difference of the Ladder Network compared to most of the other methods is the joint use of the labeled and unlabeled data. This has proven to be effective in the case of only a few labeled examples. In the following section, we investigate the effect of different noise levels, injected into the layers, as well as the effect of PCA as a preprocessing step and different denoising costs for the reconstruction of the layers.

\begin{figure}[!htb]
\begin{center}
\includegraphics[width=3.5in]{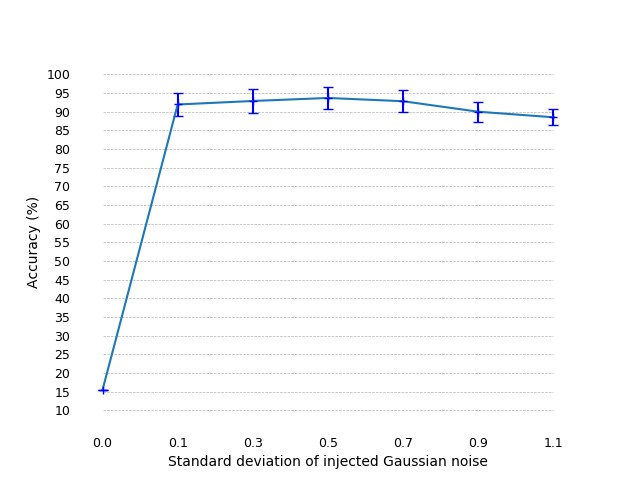}
\caption{Accuracy of the convolutional Ladder Network with different standard deviations for the injected noise.}
\label{fig6}
\end{center}
\end{figure}

Figure 6 illustrates the effect of different standard deviations of noise injected into the layers of the encoder. The noise encourages the learning of a denoising function that must learn to extract valuable information. However, denoising is only applied when a poor reconstruction induces some cost. That is why the Ladder Network has a denoising cost for each layer. The effect of having no reconstruction cost can be seen in Figure 7. For the experiment in Figure 6, we used a constant denoising cost for all layers, namely 0.1. This allows us to show the effect of different noise levels, without the side-effect of unevenly distributed denoising costs. As can be seen, no noise (std =0.0) causes the Ladder Network to have utterly poor performance. The injection of minor noise lifts up the accuracy level dramatically and the best standard deviation lies around 0.5. Note that too much noise causes a high level of corruption in the input, making it nearly impossible to extract features, as can be seen towards the end of the line.

\begin{figure}[!htb]
\begin{center}
\includegraphics[width=3.5in]{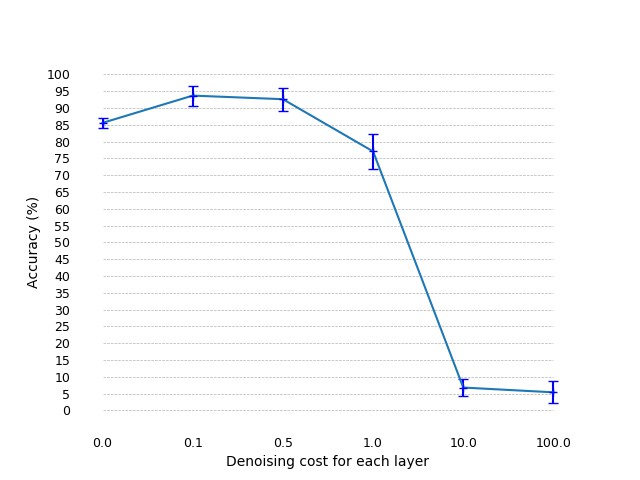}
\caption{Accuracy of the convolutional Ladder Network with different reconstruction costs at constant noise injection level (0.5).}
\label{fig7}
\end{center}
\end{figure}

Figure 7 shows that in the case of only few labeled data points, the reconstruction cost is crucial for learning meaningful features. However, one has to tune this parameter with caution, since too much emphasis on the reconstruction causes extraction of features irrelevant to the supervised task, as can be seen in Figure 7. Because the chance of picking the wrong features for reconstruction is partially eliminated by dimensionality reduction, we applied PCA as a preprocessing step. Figure 8 confirms the hypothesis that training on too many principle components leads to overfitting on irrelevant features and training on too few principle components causes loss of valuable information. Just like the noise level, something in between is the optimal choice for above reasons.

\begin{figure}[!htb]
\begin{center}
\includegraphics[width=3.5in]{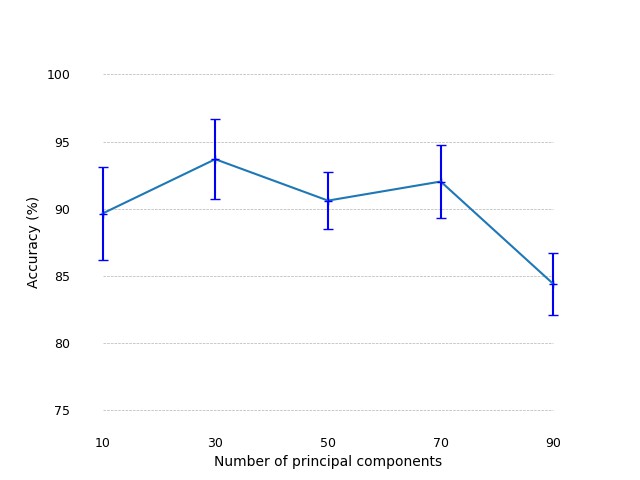}
\caption{Accuracy of the convolutional Ladder Network with different number of principle components for training.}
\label{fig8}
\end{center}
\end{figure}

\section{Discussion}
We have shown that the Ladder Network is a highly competitive architecture for semi-supervised hyperspectral image classification. In particular, we have shown that the performance of the Ladder Network given only very few labeled data points per class is superior to all state-of-the-art methods. However, there still remain some problems with this architecture: (1) The convergence, given only few labeled datapoints, is not guaranteed and therefore causes high fluctuations in the accuracies compared to the fully-labeled case. (2) The temporal-like structure of the spectral bands in each datapoint is not being exploited. \cite{7914752} have successfully applied deep Recurrent Networks (RNN) to hyperspectral image classification by making use of the sequence-like structure of the data. In ''Recurrent Ladder Networks'', \cite{DBLP:journals/corr/IlinPHRBV17}  propose a recurrent extension of the Ladder Network. However, as for the convolutional Ladder Network, the implementation is far from trivial and we are currently working on applying it to HSI classification.

\section{Code}
The code for the Autoencoders, the fully connected Ladder Network and the convolutional Ladder Network can be found here: [$\textbf{https://github.com/jubueche/Convolutional-LadderNet}$].

\clearpage

\footnotesize
\bibliographystyle{apalike}

\end{document}